%% file: paper.tex
\documentclass[a4paper,twoside]{article}

\usepackage{epsfig}
\usepackage{subcaption}
\usepackage{calc}
\usepackage{amssymb}
\usepackage{amstext}
\usepackage{amsmath}
\usepackage{amsthm}
\usepackage{multicol}
\usepackage{pslatex}
\usepackage{apalike}

\usepackage{graphicx}
\usepackage{xcolor}
\definecolor{blue-violet}{rgb}{0.54, 0.17, 0.89}

\usepackage{soul}

\usepackage{SCITEPRESS}     % Please add other packages that you may need BEFORE the SCITEPRESS.sty package.

\begin{document}

\title{Deep set conditioned latent representations for action recognition.}

% \author{Anonymous submission}
\author{\authorname{Akash Singh\sup{1}, Tom De Schepper\sup{1}, Kevin Mets\sup{1}, Peter Hellinckx\sup{2}, José Oramas\sup{1} and Steven Latré\sup{1}}
\affiliation{University of Antwerp - imec\\
\sup{1}IDLab - Department of Computer Science\\
\sup{2}IDLab - Faculty of Applied Engineering\\
Sint-Pietersvliet 7, 2000 Antwerp, Belgium}
\email{\{akash.singh, tom.deschepper, kevin.mets, peter.hellinckx, jose.oramas, steven.latre\}@uantwerpen.be}
}
\keywords{Action Recognition, Deep Sets, Deep Learning}

\abstract{In recent years multi-label, multi-class video action recognition has gained significant popularity.
While reasoning over temporally connected atomic actions is mundane for intelligent species, standard artificial neural networks (ANN) still struggle to classify them.
In the real world, atomic actions often temporally connect to form more complex composite actions.
The challenge lies in recognising composite action of varying durations while other distinct composite or atomic actions occur in the background.
Drawing upon the success of relational networks, we propose methods that learn to reason over the semantic concept of objects and actions.
We empirically show how ANNs benefit from pretraining, relational inductive biases and unordered set-based latent representations.
In this paper we propose deep set conditioned I3D (SCI3D), a two stream relational network that employs latent representation of state and visual representation for reasoning over events and actions.  
They learn to reason about temporally connected actions in order to identify all of them in the video.
The proposed method achieves an improvement of around 1.49\% mAP in atomic action recognition and 17.57\% mAP in composite action recognition, over a I3D-NL baseline, on the CATER dataset.}
\vspace{-0.4cm}
\onecolumn \maketitle \normalsize \setcounter{footnote}{0} 
% \vfill
\vspace{-0.5cm}
%%%%%%%%% BODY TEXT
\input{sections/01-introduction}

\input{sections/02-related_work}
\input{sections/04-method}
\input{sections/05-experiments}
\input{sections/06-discussion}
\input{sections/07-conclusion}

\bibliographystyle{apalike}
{\small
\bibliography{paper}}

\end{document}

%% file: sections/01-introduction.tex
\section{\uppercase{Introduction}}
\label{sec:introduction}
\begin{figure}
    \centering\includegraphics[width=1\linewidth, height=3.5cm]{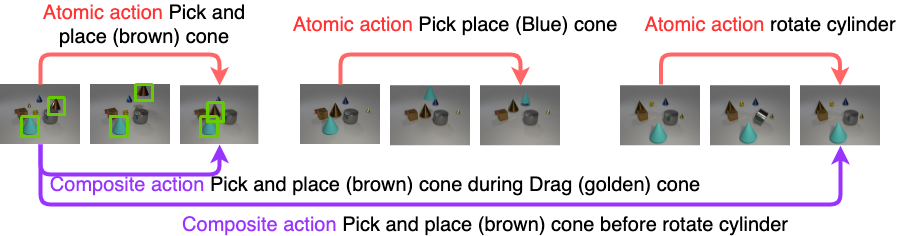}
    \caption{\textcolor{green}{Bounding box} shows an example of a change in spatial relationships between objects.
    The figure also shows how \textcolor{red}{Atomic actions} temporally relate to form \textcolor{blue-violet}{Composite actions}.
    \textcolor{blue-violet}{Composite action of }\texttt{Pick and place (brown) cone before rotate cylinder} is an example of temporal relation between actions.
    At any given time multiple \textcolor{red}{Atomic actions} and \textcolor{blue-violet}{Composite actions} may occur. 
    While \textcolor{red}{Atomic actions} are of fixed length in time, \textcolor{blue-violet}{Composite actions} can be of variable duration. } 
    \label{fig:Intro_cater}
    \vspace{-0.2cm}
\end{figure}

\vspace{-0.4cm}
% An image is worth a thousand words, does that make a video worth ten thousand?
Videos extend the semantic information of images in the temporal domain like natural language.
The series of temporal and spatial changes in videos are commonly called \textit{events}; events temporally connect in a structured manner to form \textit{atomic actions}, which at the same time combine themselves to form \textit{composite actions} \cite{girdhar2020acCATERDiagnosticDataset2020,zhu2020acComprehensiveStudyDeep2020}.
For example, in the composite action of \texttt{putting down a glass after drinking water},
\texttt{drinking} and \texttt{putting down} are atomic actions and \texttt{after} is a temporal connection between them.
%
% A specific case of an action spanning over the whole video is called a \textit{trimmed action} \cite{hutchinson2020acVideoActionUnderstanding2020}.
% Humans have the ability to understand videos by reasoning about spatial and temporal changes.
% Humans weave the event-based information to recognize an action, and extend this knowledge to recognize how actions may temporally connect and relate.
Humans understand, recall memories and objects in an unordered fashion~\cite{holtgraves1990ordered}.
We can reason about temporally connected actions also reason about objects, their attributes and relation between objects involved in actions.
Spatial and temporal understanding of events, actions and objects play an important role in tasks like action recognition, action prediction, human-object interaction etc.
While temporal and spatial reasoning is natural for intelligent species, standard artificial neural networks (ANN)
do not inherently have this ability.
For complex and human-like spatio-temporal reasoning, an ANN not only should comprehend the concept of objects, their relations but also how events and actions temporally relate as well.
For example, \texttt{pick and place cone} contains temporally related events where the spatial relation of the cone changes with respect to the table and other objects.
\texttt{Pick and place cone before rotate cylinder} additionally contains temporally related actions (Fig.\ref{fig:Intro_cater}).

\vspace{-0.1cm}
The other major challenge in action recognition in general, is that actions can take place anywhere along the time dimension. Thus, lacking a clear start and end point.
This introduces the requirement of additional reasoning related to the duration of an action, which incurs a high computational cost ~\cite{bobick1997movement,hutchinson2020acVideoActionUnderstanding2020,shoham1987reasoning,zhu2020acComprehensiveStudyDeep2020}.
Multi-label and multi-class action recognition further adds to the challenge as 
the method needs to recognise an action while disregarding or taking into consideration multiple other actions or non-action related elements that may be occurring simultaneously~\cite{hutchinson2020acVideoActionUnderstanding2020}.
While methods like I3D performs well on datasets like HMDB51~\cite{Kuehne11} and UCF101~\cite{soomro2012ucf101}, recent
studies~\cite{girdhar2020acCATERDiagnosticDataset2020,10.1007/978-3-319-49409-8_2} show how actions are highly co-related to scene bias in the above-mentioned datasets .
For example classifying \texttt{playing} a given sport based on a playfield always occurring in the background.

%While standard ANNs lack the reasoning possessed by humans they have shown high performance in finding, representing and exploiting patterns in data.
%
Recent studies~\cite{hu2018acRelationNetworksObject2018,santoro2017acSimpleNeuralNetwork2017,shanahan2020acsExplicitlyRelationalNeural2020} show that if provided with additional relational data while training, ANNs learn to represent and perform better in complex tasks like object detection and Visual QA as well. Relational networks further influence other ANN layers for relational reasoning.
%
%\cite{santoro2017acSimpleNeuralNetwork2017} in their work
%further dive deep to show how relational networks influence other ANN layers to produce implicit object-like representations for relational reasoning.
%

In our study, we take advantage of the above mentioned forte of ANNs.
We build our work on the relational network conditioned ResNet50 for action recognition~\cite{NEURIPS2019_6e79ed05}.
The ResNet50 was trained conjointly with relational network for \textit{objects state prediction task} (colour, shape, size, position) of the deep set prediction network \cite{NEURIPS2019_6e79ed05}.
The relational network and ResNet50 were optimised using mean square error and set loss during the training of deep set prediction network to output the same latent representation.
Inspired by I3D~\cite{carreira2017quo}, we extend the deep set conditioned ResNet from 2D to 3D to reason about change in the state of objects.

% In this paper we compare and study a subset of methods that span on topics of initialisation (ImageNet, CLEVR), 3D CNN, 3D CNN with LSTM for the CATER dataset. \cite{deng2009imagenet,Johnson_2017_CVPR}.
%
We propose SCI3D, a class of methods inspired by I3D \cite{carreira2017quo}, two-stream network~\cite{simonyan2014two} and Non-local neural network~\cite{wang2018non} for action recognition.
We explore the change in the states of objects representations, visual representations and space-time relation between representations.
We refer to the inflated 3D ResNet50 as \textit{DSPN} for rest of the study.
%
%
% In the study we inflate the pretrained ResNet50 from 2D to 3D  to explore the effect of change in state of objects and their relations for variable duration composite action recognition.
%
% This study explores the effect of reasoning about semantic objects and the relationships among them may have on the variable duration composite action recognition task.
% The method learns to recognise actions over untrimmed videos.

To show the effectiveness of our spatiotemporal relational methods, we chose atomic and compositional action recognition tasks offered by the CATER dataset (Sec.\ref{sec:dataset})~\cite{girdhar2020acCATERDiagnosticDataset2020}.
Unlike popular dataset, the CATER dataset minimises scene biases~\cite{girdhar2020acCATERDiagnosticDataset2020,carreira2017quo,wang2016temporal}.
%
% The dataset was built over the CLEVR image dataset~\cite{Johnson_2017_CVPR}, aims to minimise the scene bias in the video dataset domain (Sec.\ref{sec:dataset}).
%
Here SCI3D outperforms the baseline, i.e. R3D-NL \cite{wang2018non}, by 1.49\% and 17.57\% mAP in atomic and composite action recognition, respectively.
%

% To understand actions there have been multiple methods that have been proposed, however literature agrees how these methods are difficult to do fair comparison. (ref 2012.06567) 

The technical contributions of this work are
\begin{itemize}
    \item We propose a relational learning formulation over events, actions that takes in consideration sets of objects and pixels.

    \item The proposed methods are capable of generalising better for actions of variable duration on trimmed and untrimmed videos.
    % \item The proposed methods also generalise better for variable duration actions. 
\end{itemize}
\vspace{-0.8cm}
% that we propose a relational learning formulation over events, objects and propose a new deep convolutional neural network architecture.
% %
% The proposed methods are capable of generalising towards both trimmed and untrimmed multi-action recognition.
% %
% The methods also generalise for variable duration actions.
% %
% Inspired from the non-local relational networks, the proposed work also explores the relations in space, time in other words space-time domain for intra stream and inter stream \cite{wang2018non}.
%
% Stream one is initialized from an inflated set-based latent representation of objects trained using the deep-set prediction network (DSPN) \cite{NEURIPS2019_6e79ed05}.
% Stream two is I3D, which is a 3D convolutional neural network (CNN), initialized from ResNets \cite{He_2016_CVPR}. 
%

%% file: sections/02-related_work.tex
\section{\uppercase{Related work}}
\label{sec:related_work}
\vspace{-0.2cm}

\begin{figure}
    \centering\includegraphics[width=0.8\linewidth, height=5cm]{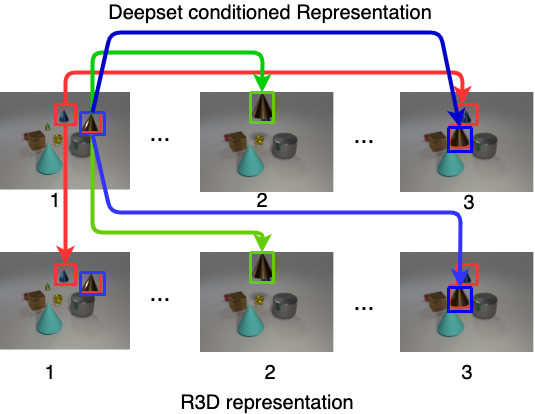}
    \caption{The relational module takes into consideration space, time, space-time relations between objects. Considering Deep-set representations on frames labelled 1, 2, 3. \textcolor{red}{Red} shows how spatial relation i.e distance or direction changes between 2 different cones. 
    \textcolor{green}{Green} shows the change of spatial relation i.e position with respect to the table and other objects but is calculated for the same object it is temporal change.
    \textcolor{blue}{Blue} shows the change in spatial and temporal relations combined. The relational module calculates it inter and intra stream.}
    \label{fig:space_time_relational}
    \vspace{-0.2cm}
\end{figure}
The study in the domain of action recognition was traditionally dominated by handcrafted features \cite{FernandoAl:TPAMI16,lan2015beyond,peng2014action,wang:inria-00583818,6751553}.
However, with better understanding of CNN architectures and of transfer learning, the focus transitioned to learning the problem in a differential manner.
In this section, we summarize the work with respect to architectures based on one-stream and two-stream networks. 
We group multi-stream networks under two-stream category.

% \subsection{Single stream architecture}
With a focus on a frame to frame prediction, single-stream networks lack sensitivity to the temporal domain. \cite{hara2018can,he2019stnet,hutchinson2020acVideoActionUnderstanding2020,ji20123d,jiang2019stm,karpathy2014large,taylor2010convolutional,tran2015learning}.
The idea is to perform image recognition, where features were extracted and the output of the method was a prediction \cite{hutchinson2020acVideoActionUnderstanding2020}.
% The network performed well with tasks like \cite{monfort2019moments,karpathy2014large}.
With a lack of temporal understanding of the data, the single-stream networks were often coupled with LSTM or with new modules and blocks \cite{donahue2015long,yue2015beyond,ghadiyaram2019large,luo2019grouped,tran2018closer}.
% New modules and blocks were also introduced to incorporate the temporal information \cite{ghadiyaram2019large,luo2019grouped,tran2018closer}. 

% \subsection{Two stream architecture}
% Temporal domain consideration is an essential part of understanding the content of the video.
For Temporal domain consideration, the CNN's were often coupled with the optical flow to capture the temporal relationship between the frames. \cite{horn1981determining,zhu2020acComprehensiveStudyDeep2020,simonyan2014two}.
% The early efforts to combine spatial stream and temporal stream were from \cite{simonyan2014two}.
While with the complement of temporal data, CNN based approaches come close to outperforming (UCF 88\% vs 87.9\% \cite{soomro2012ucf101})  or outperformed (HMDB51 59.4\% vs 61.1\% \cite{Kuehne11}) handcrafted methods, yet they still needed pre-computation.
While methods like TSN \cite{wang2016temporal} try to learn to reason on the temporal domain, they still lack the capability of modelling concepts such as objects and their spatial domain.

Two-stream networks \cite{simonyan2014two} still form the cornerstone and inspiration in the video understanding domain.
Methods like MotionNet \cite{wu2020motionnet}, MARS \cite{crasto2019mars}, D3D \cite{stroud2020d3d}, Feichetenhofer et al \cite{feichtenhofer2017spatiotemporal}, Slowfast \cite{feichtenhofer2019slowfast}, take inspiration from the two-stream networks.
While two-stream networks and I3D perform action recognition on datasets like UCF101 \cite{soomro2012ucf101} , Sport1M \cite{KarpathyCVPR14}, THUMOS \cite{jain2014university}.
They still struggled in situations where underlying actions are characterised and relies on spatial and long temporal relations.
The above-mentioned methods focus on convoluting information in a very local temporal area.
The Non-local neural network~\cite{wang2018non} when combined with I3D, try consolidating long term dependency.
Nevertheless, their capabilities have not been fully utilised. Convolutional neural networks have been shown to lose this useful temporal information down in successive stages of deep neural networks \cite{zhu2020acComprehensiveStudyDeep2020}.
The limitation of I3D in temporal reasoning is more apparent with composite action cases in the dataset like \texttt{pick place (brown) cone before rotate cylinder} Fig.~\ref{fig:Intro_cater}. 
The previously mentioned composite action consists of two atomic actions namely \texttt{pick place (brown) cone} and \texttt{rotate cylinder}. 
Furthermore, multiple atomic (\texttt{Pick place (Blue) cone}) and compositional (\texttt{Pick and place (brown) cone during drag (gold) cone}) actions can occur simultaneously Fig.~\ref{fig:Intro_cater}.

Two-stream networks could be extended to multi-stream where the other streams can augment the network with more information like audio, optical flow, or a new convolution working using different hyperparameters. 
In our work we take inspiration from two-stream network as well.
% We combine I3D with an inflated, pre-trained DSPN \cite{NEURIPS2019_6e79ed05}.

Recently methods on Transformers \cite{dosovitskiy2020image,vaswani2017attention} have gained momentum thanks to the strong semantic nature of the transformers \cite{arnab2021vivit,bertasius2021space}.
% Our work differs from the previous work \cite{bertasius2021space}.
The authors use the self-attention strategy on patches of image over space, time and space-time, we focus on the semantic concept of objects and the relational nature of non-local neural networks in space-time.
\vspace{-0.5cm}

%% file: sections/04-method.tex
\section{\uppercase{Proposed Method}}
\label{sec:method}
\vspace{-0.2cm}
\textbf{Motivation}: 3D convolutions have proven suitable for the  recognition of short duration (1-5 seconds) actions resembling the atomic actions described in Task 1. (Sec.~\ref{sec:dataset:task1}) \cite{carreira2017quo,tran2015learning,tran2017convnet,wu2019long}. %
% They lie in the center of the proposed methods to extract local temporal and spatial information from videos.
%
However, 3D CNNs tend to perform poorly when employed for very long temporally connected action .

% The method can be characterised as a single network that operates on input but comprehends the input 
% heterogeneously i.e. the state of the objects and the pixels that leads to the states.
% %
% The method combines the cognitive concept of objects, the changes in the state of the objects and latent representation to recognise actions.
%
% The method uses 3D convolution to extract temporally and spatially local/(local space-time) information from the video.

% The method localises the information using 3D convolution for atomic actions (Sec.\ref{sec:dataset:task1}).
%
To be able to reason over very long temporally connected actions, i.e composite actions (Sec.~\ref{sec:dataset:task2}), we draw inspiration from humans and use LSTM and 3D convolution. 
When reasoning about past or future, humans tend to divide time in two frames, a coarser time frame to identify temporal regions of interest and a finer, more localised, frame to reason about local space-time details.
%
% For \eg trying to remember the opening scene from a movie when the movie has ended.

\begin{figure}
    \centering\includegraphics[width=0.7\linewidth, height=1.5cm]{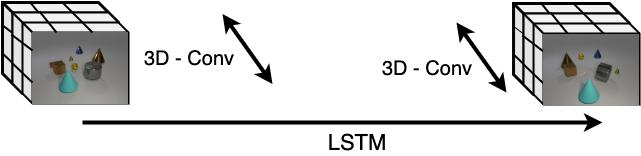}
    \caption{ Humans dissect time to reason about the past or future. 3D convolution reasons on localised temporal domain and LSTM reasons on longer temporal domain. The above figure shows how our method dissects the time to recognise composite actions of variable temporal length. }
    \label{fig:memory_disection}
    \vspace{-0.15cm}
\end{figure}

Similarly, we bifurcate time in coarse and fine frames, where 3D CNNs model  spatio-temporal details in a short temporal window and LSTMs addresses reasoning over long temporally connected action components (Fig.~\ref{fig:memory_disection}).
\vspace{-0.1cm}
% LSTM and 3D conv for the task of long temporally connected action recognition (Fig.\ref{fig:memory_disection}).

\begin{figure}[ht]
    \centering\includegraphics[width=1\linewidth, height=5cm]{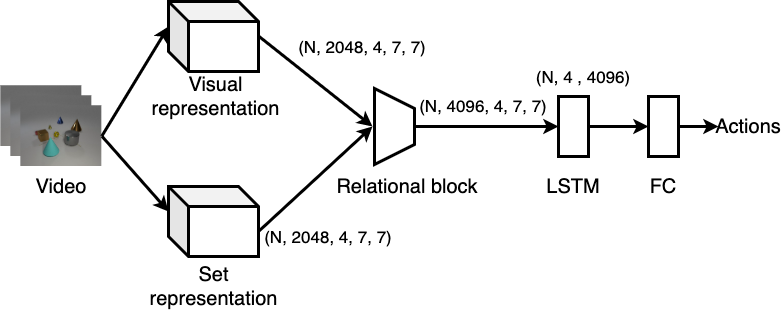}
    \caption{Proposed SCI3D architecture for action recognition.}
    \label{fig:proposedMethod}
    \vspace{-0.2cm}
\end{figure}

\subsection{SCI3D}
\label{sec:sci3d}
\vspace{-0.25cm}
We define \texttt{SCI3D} as a set-conditioned two-stream network that employs relational networks (Sec.~\ref{sec:relational_block}) to relate DSPN (Sec.~\ref{sec:dspn_pathway}) and R3D (Sec.~\ref{sec:i3d_pathway}).
The method uses non local as relational network to reason in the space-time domain over the representations (Fig.~\ref{fig:space_time_relational}).
The inspiration behind the architecture is to take advantage of the visual representation of frames and set state of the objects in the frame. Thus, augment the reasoning of R3D with DSPN (Fig.~\ref{fig:multiple:task1_two_stream}).
Both streams convolute in local space time to influence each other during training.
The other idea that forms the core of the proposed method is that we want to extend the spatial relations to temporal connected events.

With the above mentioned inspiration, we formulate SCI3D, where streams are merged using relational blocks.
The standard architecture of the proposed SCI3D is presented in Fig.~\ref{fig:proposedMethod}.
When not employing any relational block, we refer the architecture as \textit{SCI3D-NR}.
\vspace{-0.25cm}
% The single stream SCI3D is initialised only from the DSPN.

% % In ablation study (Sec.~\ref{ablation}), we also study the relational properties of non-local neural networks.
% % 3D-ResNet50 is one of the most popular backbones in action recognition \cite{wang2018non, carreira2017quo, feichtenhofer2019slowfast}.
% % Thus we limit the study of SCI3D and baselines to 3D ResNet50 backbones
% Fig.~\ref{fig:one_stream_dspn} presents the architectures that forms the foundation of all baselines and one stream SCI3D.
% For Task 2 (Sec.~\ref{sec:dataset:task2}), we extend the architecture with LSTMs.

% \subsection{DSPN pathway}
\subsection{Set Representation Stream}
\label{sec:dspn_pathway}
\vspace{-0.2cm}
% In theory, the DSPN can be any convolutional model that encodes the states of an unordered set of objects.
In theory, the set representation block can be any convolutional model that encodes the states of a set of objects.
%
% We borrow, the DSPN from Zhang \etal \cite{NEURIPS2019_6e79ed05}, deep-set prediction network for state prediction trained on the CLEVR \cite{Johnson_2017_CVPR}.
%
In practice, we extend the DSPN encoder of \cite{NEURIPS2019_6e79ed05} from ResNet34 to ResNet50 and inflate it from 2D to 3D for action recognition (Sec.\ref{sec:method:model}).

The ResNet50 was pretrained cojointly with the relational network to encode the image for state prediction task.
The task was to implicitly learn which object in the image corresponds to which set element with the associated properties(x, y, z coordinates, shape, colour, size, material) \cite{NEURIPS2019_6e79ed05}.
The latent representation learned by the ResNet50, when decoded translates to objects and their properties.
When extending the architecture from 2D to 3D, we take advantage of convolution operation in temporal dimension.
The operation looks at a series of consecutive elements(frames) to detect features, in our case the change in position as embedded in the latent space.
\vspace{-0.2cm}
%
% The ResNet50 was pretrained conjointly with relational network to represent a set of objects (position, shape, colour size etc) in a given image.
%
% The motivation to inflate the ResNet comes from the fact that 3D CNNs model temporal information better in a local neighbourhood.
%
% They will take advantage of the change in the state of the objects between the frames for the action recognition.
%

% \subsection{R3D pathway}
\subsection{Visual Representation Stream}
\label{sec:i3d_pathway}
\vspace{-0.2cm}
Visual representations from input frames are encoded via 3D CNNs along its corresponding pathway . In practice, we adopt a similar block to I3D~\cite{carreira2017quo} which takes advantage of stacked 3D CNNs and residual connections for spatio-temporal reasoning.
% introduced by Carreira \etal , 
% uses stacked 3D CNNs that take advantage of residual connections for spatio-temporal reasoning.
%
~\cite{carreira2017quo} inflate the ImageNet pre-trained 2D CNN to 3D by adopting work from~\cite{wang2015towards,zhu2020acComprehensiveStudyDeep2020}.
This architecture, i.e. I3D, is often referred to as R3D, when initialised from ResNet~\cite{girdhar2020acCATERDiagnosticDataset2020,10.1007/978-3-319-49409-8_2,wang2018non}
\vspace{-0.2cm}
\subsection{Relational Block}
\label{sec:relational_block}
\vspace{-0.2cm}
Relational networks are subsets of neural networks that embed structure with relational reasoning.
The idea is to capture the explicit or implicit relations embedded in the data.
As introduced in ~\cite{santoro2017acSimpleNeuralNetwork2017} relational network can be expressed as:
\begin{equation}
   \mathrm{RN}(O)=f_{\phi}\left(\sum_{i, j} g_{\theta}\left(o_{i}, o_{j}\right)\right)
   \label{eq:relational_eq}
\end{equation}

where the input is a set that can be expressed as an abstract humane concept. 
It can be pixels, features \cite{wang2018non}, entities, objects \cite{santoro2017acSimpleNeuralNetwork2017} or frames \cite{zhou2018temporal}.
In our formulation, $O$ is defined by the input video, $o_{i}, o_{j}$ are the outputs from the two streams/pathways, 
whereas   $ f_{\phi} $ and $g_{\theta}$ are functions to relate the outputs.

% In the proposed method, we try to bring lateral representation of states and vision to reason on space-time domain.
% In our work, we try to bring the best of both the human and machine world, i.e cognitive concept of object and reasoning over pixels.
%
The relational block reasons about an event or an action on latent representation of states and of visual in space-time domain.

We employ the non-local neural networks as a relational block, which given a position they compute the weighted sum of features to all other positions as follows:

\begin{equation}
   \mathbf{y}_{i}=\sum_{j \in \Omega} \omega\left(\mathbf{x}_{i}, \mathbf{x}_{j}\right) g\left(\mathbf{x}_{j}\right) 
   \label{eq:non_local_eq}
\end{equation}

Where $ \mathbf{x}_i$ represents a feature at position $i$, $ \mathbf{y}_i$ is the output tensor. 
$\omega$ is similarity function between $i$ and $j$, in our case we evaluated dot-product, gaussian and embedded gaussian. $g(\mathbf{x}_j)$ is the pixel representation at point $j$.
The non-local block performs relational computations, analogous to relation networks \cite{battaglia2018relational,levi2018efficient,yin2020disentangled,zambaldi2018relational}.
Non local neural networks can be considered a set to set architecture, 
where they expect as input a set of features and output the transformed set of features.
\vspace{-0.67cm}
% The authors of the CATER dataset \cite{girdhar2020acCATERDiagnosticDataset2020} also benchmark the performance on R3D, to present the utility and need of CATER like video datasets with respect other datasets like Kinetic, UCF-101 and HMDB-51 \cite{carreira2017quo, Kuehne11, soomro2012ucf101}.
%

%% file: sections/05-experiments.tex
% \section{Experiments: Action recognition}

\section{\uppercase{Evaluation}}
\label{sec:experiment}
\vspace{-0.3cm}
\input{sections/03-Dataset.tex}

\subsection{Implementation Details}
\label{sec:method:model}
\vspace{-0.3cm}
In this section, we define the implementation details of SCI3D.

% \subsubsection{DSPN pathway}
\textbf{DSPN}. For the implementation, we extend the backbone of~\cite{NEURIPS2019_6e79ed05} from ResNet34 to ResNet50 to train our DSPN backbone. 
The $3\times3$ kernel in a residual block of ResNet50 is inflated to $3\times1\times1$, as discussed by~\cite{feichtenhofer2016spatiotemporal} and~\cite{wang2018non}.
As suggested by~\cite{wang2018non}, we also constrain the computation by inflating only one kernel for every two residual blocks.
Apart from lowering down the number of computations, the above-mentioned inflation strategy also leads to better results \cite{wang2018non}.
%
% For atomic action recognition (Sec.~\ref{sec:dataset:task1}), the single stream SCI3D is supplemented with one FC but for longer composite action recognition (Sec.~\ref{sec:dataset:task2}) we supplement it with 2 LSTM and 1 FC.

% \subsubsection{I3D pathway}
\textbf{R3D/R3D-NL}. For the implementation, we again follow the inflation details from~\cite{wang2018non}. 
We initialize the weights with pretrained ResNet50 weights.
Similar to the DSPN, the $3\times3$ kernel is inflated to $3\times1\times1$.
Otherwise mentioned explicitly, all other details of the architecture is followed as discussed in~\cite{wang2018non}.

\textbf{SCI3D}
For SCI3D (Fig.\ref{fig:proposedMethod}), we employ relational block (Sec.\ref{sec:relational_block}) to combine DSPN pathway (Sec.\ref{sec:dspn_pathway}) and R3D pathway (Sec.\ref{sec:i3d_pathway}).
For shorter atomic action recognition (Sec.\ref{sec:exp:task1}), the LSTM component was redundant in SCI3D (Fig.\ref{fig:proposedMethod}), thus we remove it to only use fully connected layer.
\vspace{-0.2cm}
\subsection{Training Details} 
\vspace{-0.2cm}
All of the experiments were performed on 2 NVIDIA V100 GPUs.
We adopted baseline LR to 0.0025 according to the linear scaling rule \cite{goyal2017accurate}. LR, for our methods in Task 1 and Task 2 were 0.015 and 0.0025, respectively.
They are reduced by a factor of 10 at epochs 90 and 100. We use momentum of 0.9.
We fine-tune our method with 32-frame input clips \cite{girdhar2020acCATERDiagnosticDataset2020}.
The spatial input size is 224×224 pixels, randomly cropped from a scaled video whose shorter side is randomly sampled in [256, 320] pixels.
For our methods on Task 1 and 2, we train them for 120 and 140 epochs respectively.
%
% The learning rates for our methods in Task 1 and Task 2 were 0.015 and 0.0025, respectively.
%
% They are reduced by a factor of 10 at epochs 90 and 100. We use momentum of 0.9.
%
A dropout of 0.5 is applied after the global pooling layer and in LSTMs \cite{hinton2012improving}, weight initialization was adopted from the original work of the non-local neural networks \cite{wang2018non,he2015delving}.

\begin{table}[t]
\caption{Comparing the best performing SCI3D variant with baseline and other standard architectures.}
\begin{center}
\resizebox{\columnwidth}{!}{
\begin{tabular}{|l|c|c|c|c|}
\hline
\textbf{Task} &
\textbf{Frozen} &
\textbf{LSTM} &
\textbf{Achitecture Name} &
\textbf{mAP (\%)} \\
\hline\hline
Task 1 &  &  & latent-FasterRCNN & 63.85 \\
Task 1 & \checkmark &  &Single stream SCI3D & 69.21 \\
Task 1 &  &  & Single stream SCI3D & 91.82\\
\hline
% Task 1 &  &  & R3D-NL(R) & 98.8 \\
Task 1 &  &  & R3D-NL (\cite{wang2018non}) & 95.28 \\
Task 1 &  &  & SCI3D-NR  & 95.95 \\
Task 1 &  &  & SCI3D & \textbf{96.77}\\
\hline
\hline
Task 2 &  & \checkmark & latent-FasterRCNN & 25.45 \\
Task 2 & \checkmark & \checkmark &Single stream SCI3D & 26.32 \\
Task 2 &  & \checkmark & Single stream SCI3D & \textbf{69.76}\\
\hline
% Task 2 &  & \checkmark & R3D-NL (R) & 53.1 \\
Task 2 &  & \checkmark & R3D-NL (\cite{wang2018non}) & 52.19 \\
Task 2 &  & \checkmark & SCI3D-NR & 66.71 \\
Task 2 &  & \checkmark &  SCI3D & 65.92 \\
% Task 2 &  & \checkmark & Single stream SCI3D & \textbf{69.76}\\
\hline
\end{tabular}}
\end{center}
\vspace{-0.3cm}
\label{tab:result_table}
\end{table}

We assume a broader definition of actions, that considers the actions of both animate and inanimate actors \cite{hutchinson2020acVideoActionUnderstanding2020}.
Thus, we define an action as temporally connected events or other actions that can be of any length in time when weaved.
Action recognition is the classification of such actions.

Considering the variable duration and the broader definition of actions, the experiments with the proposed methods present the empirical results (Table.\ref{tab:result_table}) supporting how the cognitive concept of objects aids with action recognition.
\vspace{-0.2cm}
%
%Later in the section, we show an ablation study with respect to different relational modules like the dot product, gaussian and embedded gaussian.
%
\subsection{Experiments: Task 1}
\label{sec:exp:task1}
\vspace{-0.2cm}
An action of the atomic action recognition task can be expressed as, temporally connected events located in a local neighbourhood.
3D CNN are well suited for simpler action recognition tasks defined in a local neighbourhood.
They lie in the centre for all the baselines and the proposed methods.

We approach the task as a multi-label classification problem \cite{girdhar2020acCATERDiagnosticDataset2020}.
We evaluate the performance of all the methods with mean average precision (mAP).

\textbf{Baselines}: For the task, we employ R3D-NL~\cite{carreira2017quo,wang2018non} and \textit{latent-FasterRCNN}~\cite{ren2015faster} as baselines. On the one hand, R3D-NL provides a bottom-up visual representation. On the other hand, latent-FasterRCNN aims at exploiting a semantic-level representations learned from isolated objects in the dataset.
% We believe these two baselines will properly cover the representation spectrum.

% \begin{flushleft}
% \textbf{Single stream SCI3D}
% \end{flushleft}

% \textbf{Baseline:} 
% R3D-NL is established baseline, as discussed by \cite{girdhar2020acCATERDiagnosticDataset2020}.
%
We consider R3D-NL baseline as discussed by \cite{girdhar2020acCATERDiagnosticDataset2020}.
% The authors extend the R3D \cite{wang2018non}, by replacing conv3 and conv4 blocks in ResNet with Non-local block.
%
To establish the latent-FasterRCNN~\cite{ren2015faster} baseline, we train the method with ResNet50 as the backbone for object detection on the CLEVR dataset.
We extract the backbone of FasterRCNN and inflate the ResNet50 to R3D as discussed in Sec.~\ref{sec:method:model}.
%
% For the ease of reference, we refer it as \texttt{FR-R3D} in the results (Table.\ref{tab:result_table}).
%
The choice of establishing latent-FasterRCNN was influenced by OPNet\cite{shamsian2020learning}.
By reason, the latent representation of FasterRCNN are similar to DSPN, they both identify objects but in theory, they differ.
While FasterRCNN's goal is object detection, DSPN extends object detection to also model the state of the objects.
% The experiment forms the foundation of the work, it is designed to validate the advantages of the deep set conditioned latent representations for action recognition.
%
% The underlying idea of the experiment is to utilise the change in the state of the objects for action recognition.
%

\begin{figure}
    \centering\includegraphics[width=0.7\linewidth, height=2cm]{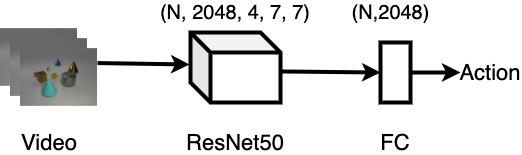}
    \caption{The architecture for single stream methods. The architecture forms the foundation for R3D-NL, SCI3D.}
    \label{fig:one_stream_dspn}
\end{figure}

We also consider the single-stream variants of SCI3D.
In these variants, SCI3D only has a single pathway which is initialised with a DSPN (see Fig.~\ref{fig:one_stream_dspn}).
We investigate the usefulness of the set conditioned latent representations with respect to latent-FasterRCNN baseline by freezing the SCI3D.
%
% For the atomic action recognition task, we compare Frozen single stream SCI3D \ref{sec:sci3d} with the above-mentioned baselines.
%
We present the results in Table \ref{tab:result_table}.

\textbf{Results}
The first observation done throughout the execution of this experiment was that given the relatively short duration of the atomic actions involved in Task 1, the LSTM component was redundant. For this reason, it was removed from the archiecture when conducting experiments related to Task 1.

\begin{figure}
    \centering\includegraphics[width=0.8\linewidth, height=4.5cm]{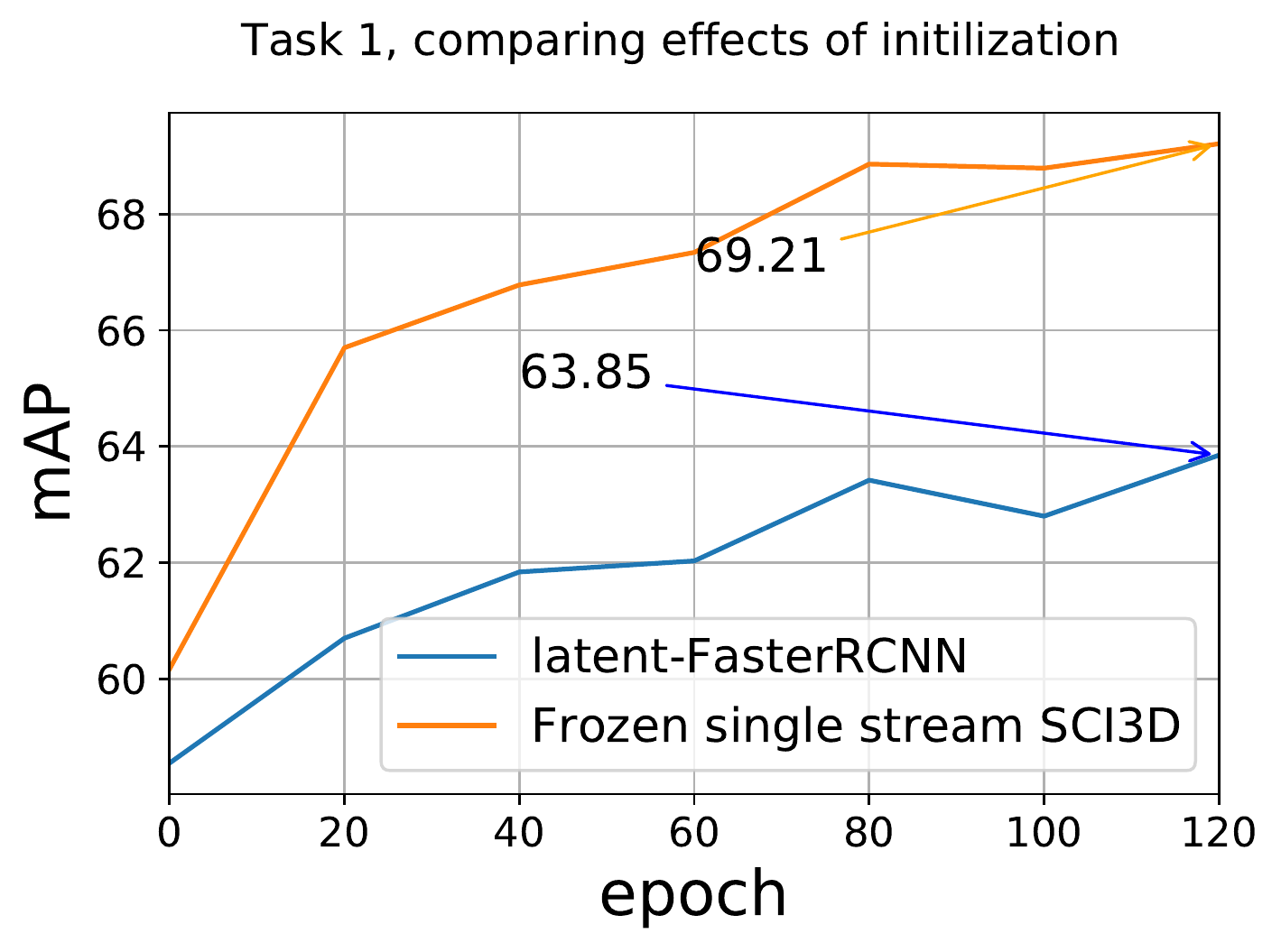}
    \caption{ Validation during training on task 1, the plot illustrates the effectiveness of weight initialization.}
    \label{fig:multiple:task1_one_stream}
\end{figure}

Regarding the single-stream baselines,
the frozen SCI3D outperforms the latent-FasterRCNN backbone by 5.36\% mAP with only FC trainable weights.
The fact that SCI3D outperforms the pre-trained latent-FasterRCNN, leads us to conclude that reasoning about set-level properties (beyond that of individual objects as done by latent-FasterRCNN) leads to the better results (Fig.\ref{fig:multiple:task1_one_stream}). Thus, supporting the benefits of the proposed conditioning on deep set-level representations . 
Yet, from Table~\ref{tab:result_table} it is clear that these single-stream are unable to outperform the state-of-the-art R3D-NL.

% Higher performance of SCI3D could be attributed to the transfer learning as DSPN was pretrained on the CLEVR dataset.
%

\begin{figure}
    \centering\includegraphics[width=0.8\linewidth, height=4.5cm]{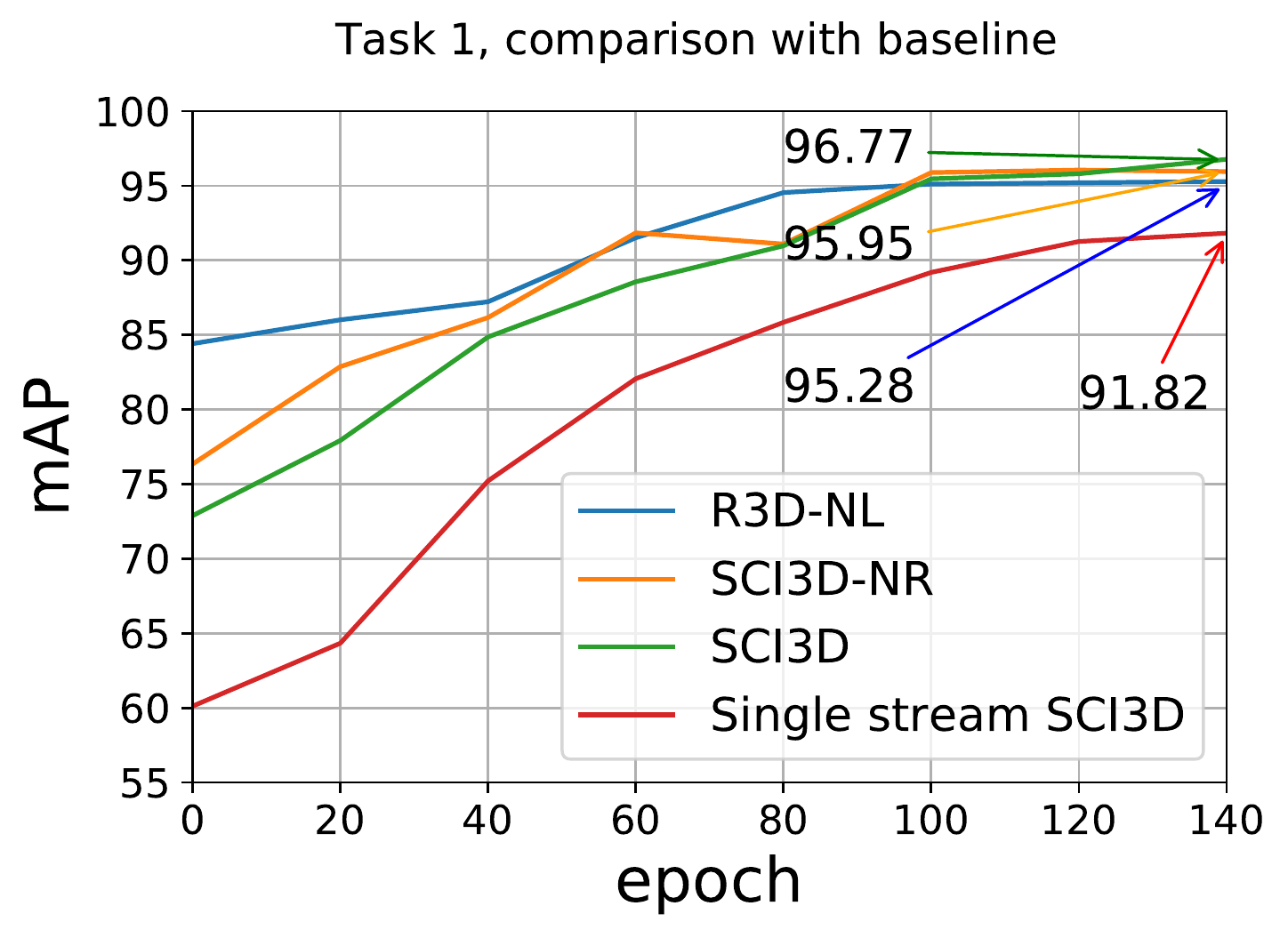}
    \caption{validation during training plot comparing different methods on task 1.}
    \label{fig:multiple:task1_two_stream}
    \vspace{-0.2cm}
\end{figure}

The last observation from above changes when we look at the proposed two-stream SCI3D. We notice that while the non-relational SCI3D-NR variant is on part with the R3D-NL baseline (mAP around 95.28\%), its relational variant outperforms it by around 1.5\% mAP (Fig.\ref{fig:multiple:task1_two_stream}).

We attribute the higher performance of R3D-NL compared to single stream SCI3D to the fact that Task 1 (Sec.\ref{sec:dataset:task1}) is an event-centric task, where events last only for a fraction of an action.
Suggesting that methods need to take in account the minor change in pixels.
The non-local configuration in the R3D-NL variant proposed by~\cite{girdhar2020acCATERDiagnosticDataset2020} is better suited to detect the change in transformations.
Moreover, the reduced difference between SCI3D (96.77\% mAP) and SCI3D-NR (95.95\% mAP) on Task 1 further strengthens our belief about the focus on very local events and pixels in Task 1.
\vspace{-0.2cm}

% \begin{flushleft}
% \textbf{SCI3D}
% \end{flushleft}
% \textbf{Baseline}: R3D-NL is established baseline, as discussed by Girdhar and Ramanan \cite{girdhar2020acCATERDiagnosticDataset2020}.
% %
% But for the R3D-NL baseline, Girdhar and Ramanan \cite{girdhar2020acCATERDiagnosticDataset2020} extend the R3D implementation from wang \etal \cite{wang2018non}, by replacing conv3 and conv4 blocks in ResNet with Non-local block.

% %
% % The outputs from the networks are merged to serve as input to an FC which classifies the actions.
% %
% %
% SCI3D logs 96.77\% mAP and SCI3D-NR logs 95.95\% mAP when compared to 95.28\% mAP by R3D-NL on Task 1
% The SCI3D performs the best further strengthening our belief about the focus on events and pixels in task 1.
% %

\subsection{Experiments: Task 2}
\label{sec:exp:task2}
\vspace{-0.25cm}
\begin{figure}
    \centering\includegraphics[width=0.8\linewidth, height=4.5cm]{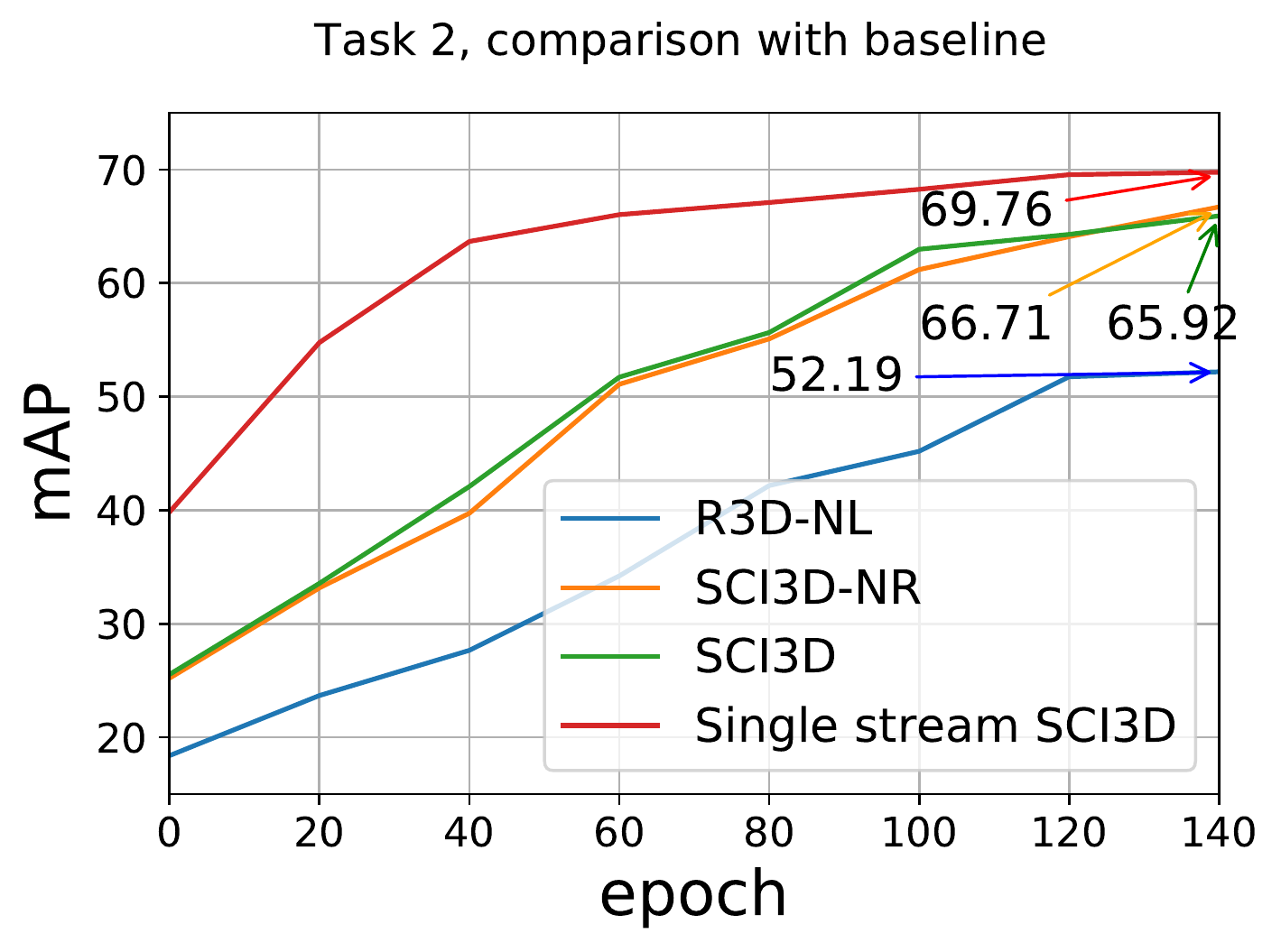}
    \caption{ Validation during training plot comparing different methods on task 2.}
    \label{fig:multiple:task2_two_stream}
    \vspace{-0.2cm}
\end{figure}

% Task 2 is an untrimmed action recognition task coined as a compositional action recognition task.
%
Task 2 extends the atomic action recognition task where 2 atomic actions temporally connect to form a composite action.
Task 2 (Sec.~\ref{sec:dataset:task2}) is inherently different from Task 1, an action commenced at frame=0 can last till the end of the video.
The methods in Task 2 need to reason for a flexible temporal range. They also need to take into account other atomic actions and composite actions occurring simultaneously. 
Thus we employ the original proposed architecture of the SCI3D and SCI3D-NR (Fig.\ref{fig:proposedMethod}).
We also extend the single-stream SCI3D (Fig.\ref{fig:one_stream_dspn}) variants from Task 1 (Sec.~\ref{sec:exp:task1}) with 2 layer LSTM.
The LSTMs in the architecture assists in longer variable length temporal reasoning.

% However, the composite action is not time-constrained i.e can last for any length of the video, employing LSTM extends the capabilities of 3D CNN architecture from local temporal reasoning to longer temporal reasoning.

% \begin{flushleft}
% \textbf{SCI3D}
% \end{flushleft}
%
%
% The CATER dataset offers 301 distinct classes for the task.
%
We approach the problem as multi-label classification, an use mAP as performance metric.

\textbf{Baselines} we follow \cite{girdhar2020acCATERDiagnosticDataset2020} where R3D-NL is extended using 2 layer LSTM with 512 hidden units.
A similar extension was applied to the latent-FasterRCNN baseline.

\textbf{Results}
\begin{table*}[!h]
\caption{Ablation study to understand the 
advantage of different building blocks, methods and their impact on respective task. Conv refers to a single convolution of kernel 3, stride 1 and padding 0. }
% \vspace{-0.35cm}
\begin{center}
\begin{tabular}{|l|c|c|c|c|c|c|}
\hline
\textbf{Task} & 
\textbf{Frozen} &
\textbf{Relational block} &
\textbf{LSTM} &
\textbf{Architecture} & 
\textbf{block}&
\textbf{mAP}(\%) \\
\hline\hline
Task 1 & \checkmark & - & - & Single stream SCI3D & 2 FC & 69.25 \\
Task 1 & \checkmark & - & - & Single stream SCI3D & 1 Conv & 68.45 \\
% Task 1 & & & & Single stream SCI3D & BasicBlock-NL & 92.42\\
% Task 1 & & & & Single stream SCI3D & BottleNeck-NL & 93.65\\
Task 1 & - & - & \checkmark & Single stream SCI3D & 256 LSTM & 91.79\\
Task 1 & - & Embedded gaussian & - & SCI3D & - & 96.17\\
Task 1 & - & Gaussian & - & SCI3D & - & 95.88 \\
\hline
Task 2 & - & Embedded gaussian & - & SCI3D & - & 65.31\\
Task 2 & - & Gaussian & - & SCI3D & - & 64.99\\
Task 2 & - & - & \checkmark & SCI3D & FC + 512 LSTM & 31.11\\
Task 2 & - & \checkmark & - & SCI3D & 512 LSTM & 53.71 \\
Task 2 & - & - & \checkmark & SCI3D-NR & 1FC + 512 LSTM & 32.35\\
Task 2 & - & - & \checkmark & SCI3D & 512 LSTM & 55.68\\
\hline
\end{tabular}
\end{center}
\label{tab:ablationTable}
\vspace{-0.6cm}
\end{table*}

At first sight the absolute performance values on this task are relatively lower compared to those on Task 1. This clearly indicated the increased complexity of this task.

We notice that the single-stream SCI3D achieves 69.76\% mAP on the task, outperforming the R3D-NL  baseline by 17.57\% mAP (Fig.\ref{fig:multiple:task2_two_stream}).
It is noticeable from training Fig.~\ref{fig:multiple:task2_two_stream} that single stream SCI3D trains faster and more efficiently.  

SCI3D and SCI3D-NR achieve 65.92\% and 66.71\% mAP (Table\ref{tab:result_table}).
They outperform baseline by 13.73\% and 14.52\% mAP respectively.
Single stream SCI3D outperforms SCI3D and SCI3D-NR by 3.84\% and 3.05\% mAP respectively.
As discussed previously, the R3D block when combined with the DSPN using the relational block in SCI3D promotes the focus on local events and shorter actions.
~\cite{wu2019long} empirically show, the relational block performs the best when combined with longer temporal representations.
\vspace{-0.2cm}
%

% Tom De Schepper <Tom.DeSchepper@uantwerpen.be>

% \begin{figure}
%     \centering\includegraphics[width=0.7\linewidth, height=2.75cm]{figures/task2_memory_disection.png}
%     \caption{ Humans dissect time to reason about the past or future. 3D convolution reasons on localised temporal domain and LSTM reasons on longer temporal domain. The above figure shows how our method dissects the time to recognise composite actions of variable temporal length. }
%     \label{fig:memory_disection}
% \end{figure}
% %

\begin{figure}
    \centering\includegraphics[width=0.8\linewidth, height=4.5cm]{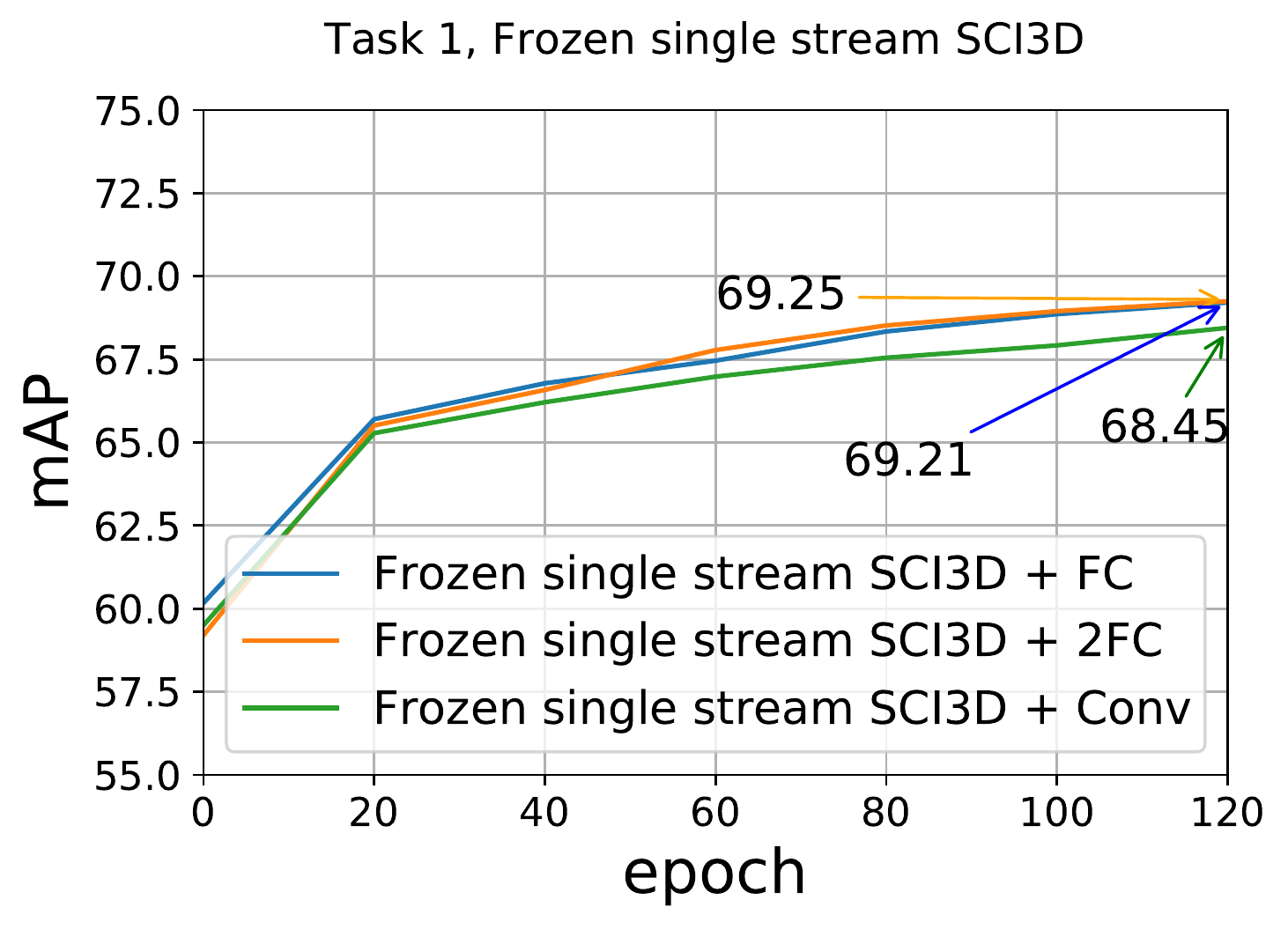}
    \caption{Validation during training plot when frozen one stream SCI3D combined with different blocks.}
    \label{fig:multiple:ablation_frozen_dspn}
    \vspace{-0.2cm}
\end{figure}

\subsection{Ablation Study}
\label{ablation}
\vspace{-0.3cm}
To fully understand the contribution of each building block, we conduct ablation studies (Table \ref{tab:ablationTable}) by adding and deleting components.

We limit the study of SCI3D and baselines to 3D ResNet50 backbones because it is one of the most popular backbones in action recognition~\cite{carreira2017quo,feichtenhofer2019slowfast,wang2018non}.
Adding more fully connected (FC) layers over frozen single stream SCI3D did not improve model performance significantly.
With 2 FCs of 2048 and 512 units each, we observed an increase of mAP of 0.04\%.
While adding a convolutional block with a FC we saw a drop in mAP of 0.8\%.
Adding more FC or convolutional layers provides little to no gain.
Using 2 LSTM layers with 256 hidden units with single stream SCI3D logs 91.79\% mAP on Task 1.

\begin{figure}
    \centering\includegraphics[width=0.8\linewidth, height=4.5cm]{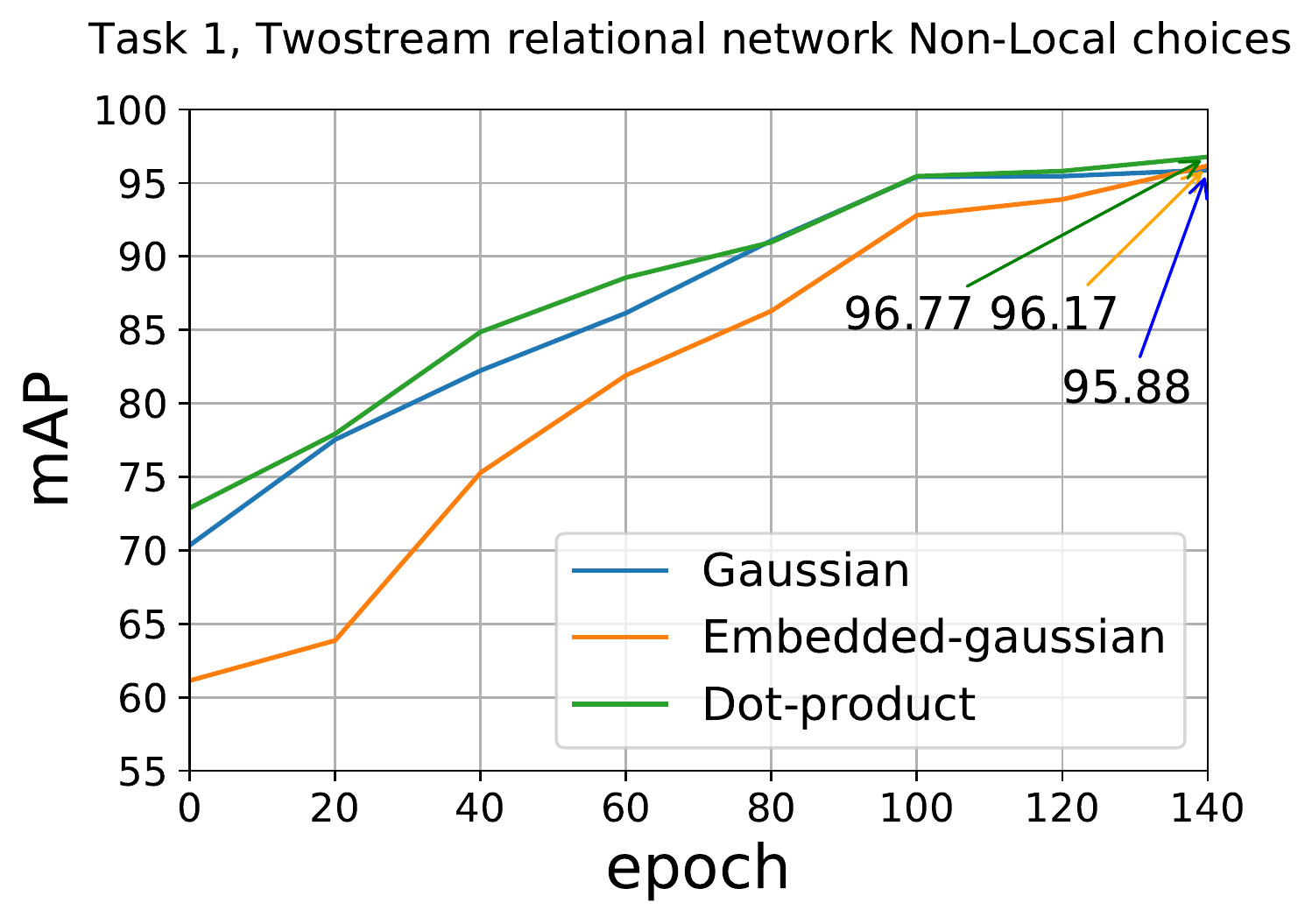}
    \caption{Ablation study of different strategies of non-local block for Task 1.}
    \label{fig:multiple:ablation_twostream_non_local_task1}
    \vspace{-0.2cm}
\end{figure}

\begin{figure}
    \centering\includegraphics[width=0.8\linewidth, height=4.5cm]{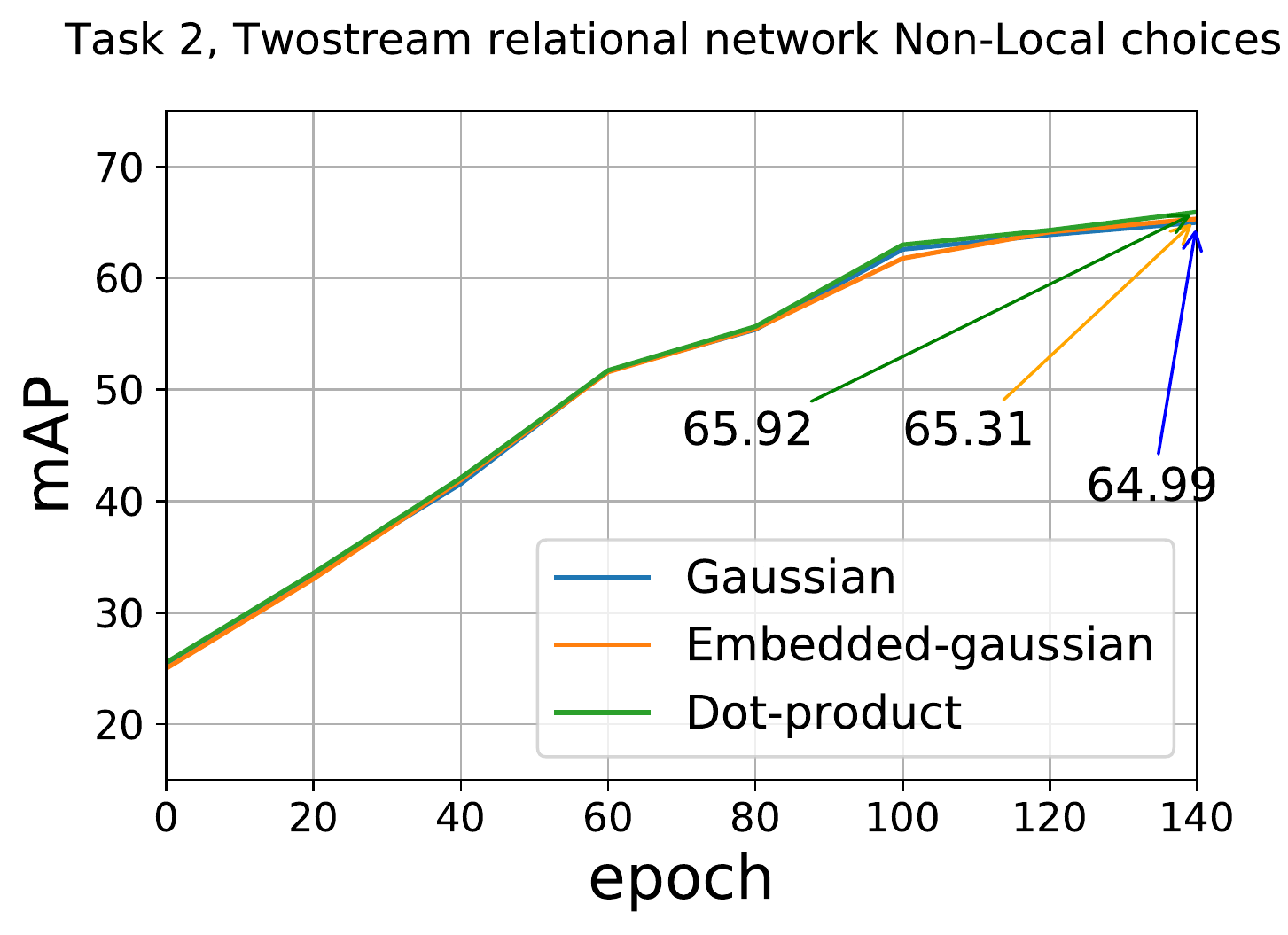}
    \caption{Ablation study of different strategies of non-local block for Task 2.}
    \label{fig:multiple:ablation_twostream_non_local_task2}
    % \vspace{-0.1cm}
\end{figure}

Addition of an additional convolutional block or an FC over SCI3D-NR architecture on Task 1 (Sec.~\ref{sec:method:model}), did not show any major impact.
The choice of the non-local strategy for the SCI3D architecture makes a little difference for Task 1, as shown in validation during training plots in Fig.~\ref{fig:multiple:ablation_twostream_non_local_task1}.
% We present the validation during training plots comparing the strategies in Fig.~\ref{fig:multiple:ablation_twostream_non_local_task1}. 
While for Task 2 there is little to no difference, see Fig.~\ref{fig:multiple:ablation_twostream_non_local_task2}.
We observe that adding an FC layer between LSTMs and the relational block flattens all temporal information and leads to under-performance.
Under this setting, the SCI3D achieves an mAP of 31.11\% on Task 2.
\vspace{-0.6cm}

% \begin{figure}
%     \centering\includegraphics[width=0.5\linewidth, height=3cm]{figures/Ablation_twostream_block_choice_task2.pdf}
%     \caption{Ablation study of different learnable blocks for two stream relational network on Task 2.}
%     \label{fig:multiple:ablation_twostream_relational_block_choice_task2}
% \end{figure}

% \begin{figure}
%     \centering\includegraphics[width=0.5\linewidth, height=3cm]{figures/Ablation_twostream_concat_block_choice_task2.pdf}
%     \caption{Ablation study of different learnable blocks for two stream concat network on Task 2.}
%     \label{fig:multiple:ablation_twostream_concat_block_choice_task2}
% \end{figure}

% \vfill

%% file: sections/03-Dataset.tex
\subsection{Dataset}
\label{sec:dataset}
\vspace{-0.2cm}
% CATER: \textbf{C}ompositional \textbf{A}ctions and \textbf{TE}mporal \textbf{R}easoning dataset \cite{girdhar2020acCATERDiagnosticDataset2020}, builds over the CLEVR dataset \cite{Johnson_2017_CVPR}, and tries to address the problem of scene bias in video dataset.
% The authors address previously mentioned shortcomings of other popular datasets \cite{girdhar2020acCATERDiagnosticDataset2020}.
%
% The authors point out how heavily biased the datasets \cite{Kuehne11, soomro2012ucf101} are by presenting evidence that state of the art pose-based action recognition models were outperformed with frame-level models.\cite{girdhar2020acCATERDiagnosticDataset2020, wang2016temporal}. 
%
% The CATER dataset, offers 3 tasks that
CATER dataset extends the CLEVR dataset \cite{Johnson_2017_CVPR}, to address the problem of scene bias in video datasets~\cite{girdhar2020acCATERDiagnosticDataset2020,wang2016temporal}.
% CLEVR is a diagnostic image dataset that aims to mimimize vision bias in visual question answering \cite{Johnson_2017_CVPR}.
We validate our method on the CATER dataset~\cite{girdhar2020acCATERDiagnosticDataset2020} that offers three tasks that focus on reasoning around cognitive concepts like causal reasoning over long term temporal structure over events.
We target their atomic action recognition (Sec.\ref{sec:dataset:task1}) and composite action recognition tasks (Sec.\ref{sec:dataset:task2}). Both multi-label classification tasks with 14 and 301 classes respectively.
% to show the reasoning and effectiveness of our methods in semantic understanding of the actions \cite{girdhar2020acCATERDiagnosticDataset2020}.
%

% The dataset supplements the challenge in form of untrimmed videos.
%
% While in a trimmed video, an action generally spans over the whole length of the video, untrimmed videos pose an additional challenge as the model needs to classify an action while disregarding other actions.

%

The dataset offers 5000 training videos and 1650 validation videos at $320\times240$ px, where a single video contains 300 frames, rendered at 24fps.
An atomic action is always constrained to a maximum of 30 frames while a composite action can last anywhere from 30 to 300 frames.
\vspace{-0.2cm}

\subsubsection{Task 1: Atomic action recognition}
\label{sec:dataset:task1}
\vspace{-0.3cm}
It is the primary action recognition task offered by the CATER dataset \cite{girdhar2020acCATERDiagnosticDataset2020}. Events temporally relate to form simple granular actions like \texttt{Pick and place cone}, \texttt{rotate cylinder} as shown in Fig.\ref{fig:Intro_cater}.

While different actions can share the same events, we believe it is a simpler of the two considered task because of the low number of classes (14) as the classification does not differentiate between object types.
The task can be extended in the future for granular event-based reasoning by extending actions classes to include object colour, size etc.
\vspace{-0.2cm}
\subsubsection{Task 2: Compositional action recognition}
\label{sec:dataset:task2}
\vspace{-0.25cm}
Real-world actions are mostly compositional in nature. 
In the composite action recognition task, the atomic actions can temporally relate in 13 categories defined in Allen's temporal algebra \cite{allen1983maintaining}.
Same as Girdhar and Ramanan~\cite{girdhar2020acCATERDiagnosticDataset2020} , we consider only 3 categories namely, \textbf{before}, \textbf{during} and \textbf{after}.
Akin to Task 1, multiple composite actions are active at any given moment in a video.
We identify that Task 2 provides us with the additional challenge that a composite action can last for a part or the whole duration of a video (Fig.\ref{fig:Intro_cater}).
From Fig.\ref{fig:Intro_cater}, an example \texttt{Pick and place cone during slide cone} may last for the same time as Task 1, while \texttt{pick and place cone before flip cylinder} actions lasts for the whole video.
% so network needs to keep that in context.
% or it can last same amount as atomic action like pick place cube during slide cone.
While the model should be capable of adapting to any temporal window to classify actions,
it should also be capable of identifying other atomic and  composite actions.
\vspace{-0.2cm}
% We approach the problem in a similar fashion as Task 1, i.e., multi-label classification over 301 classes.

% \subsection{Action recognition}
% Over last the decade there has been advancement in deep learning by leaps and bounds.
% The task of action recognition benefited from the advancement of image recognition as well.
% Although, unlike image processing, we observed new challenges in video modelling with respect to modelling of long-range time domain as well.

%% file: sections/06-discussion.tex
\section{\uppercase{Discussion}}
\label{sec:discussion}
\vspace{-0.35cm}
Action recognition from the visual appearance alone is challenging in the CATER dataset.
The dataset offers untrimmed videos, which poses an additional challenge as the methods needs to classify an action while disregarding other actions.
The study achieve an improvement of 17.57\% mAP over the baseline R3D-NL on untrimmed videos by employing the deep-set conditioned latent representation.
The latent representation embed the set of objects and attributes like shape, colour and x, y, z coordinates.
Though the study does not provide any results supporting the advantages of explicitly using the set of objects, it lays the foundation for it.
The SCI3D network under performs in Task 2 (Sec.~\ref{sec:dataset:task2}) because it is constrained by branch focused on localised events and actions.
However,~\cite{wu2019long}, show non-local relation block performs the best when supplemented with features supporting longer temporal relation.
Yet, besides this, the results show that the single-stream variant of SCI3D, which only relies on set-based representations leads the performance by a significant margin. This further strengthens the potential gains that can be achieved by shifting reasoning from individual entities to sets.
The current study can also be translated to real-world scenarios,
by modelling hands as discussed by \cite{girdhar2020acCATERDiagnosticDataset2020}.
It can also be adopted for real-world videos by adopting latent representations trained on natural images from \cite{rezatofighi2017deepsetnet}.
% \vfill
\vspace{-0.55cm}

%% file: sections/07-conclusion.tex
% \vspace{-0.1cm}
\section{\uppercase{Conclusion}}
\label{sec:conclusion}
\vspace{-0.35cm}
The study focuses on untrimmed action recognition but it generalises to recognise trimmed composite action as well.
We empirically show the advantages of deep set conditioned representations and relational networks. 
When the deep network is initialized with the representations and equipped with relational reasoning, they outperform benchmarks.
The proposed method, SCI3D outperforms the previous methods by 17.57\% mAP.
Based on the outcome and discussion of this work \cite{hu2018acRelationNetworksObject2018}, we believe set of objects and relational networks are promising components for the automatic understanding of natural videos.
While we limit our work to tasks in the CATER dataset, we believe the methods can be extended to explanatory and predictive causal reasoning tasks like why happened and what is about to happen.
% \vspace{-0.4cm}
\section*{\uppercase{Acknowledgement}}
% \vspace{-0.3cm}
This research received funding from the Flemish Government under the “Onderzoeksprogramma Artificiële Intelligentie (AI) Vlaanderen” programme.

We  would  like  to  thank MMaction2 contributors and community \cite{2020mmaction2} for tremendously helpful documentation and code. We also thank Weights and biases \cite{wandb} for free academic account to log and visualize the training.

%% file: paper.bbl
\begin{thebibliography}{}

\bibitem[Allen, 1983]{allen1983maintaining}
Allen, J.~F. (1983).
\newblock Maintaining knowledge about temporal intervals.
\newblock {\em Communications of the ACM}, 26(11):832--843.

\bibitem[Arnab et~al., 2021]{arnab2021vivit}
Arnab, A., Dehghani, M., Heigold, G., Sun, C., Lu{\v{c}}i{\'c}, M., and Schmid,
  C. (2021).
\newblock Vivit: A video vision transformer.
\newblock {\em arXiv preprint arXiv:2103.15691}.

\bibitem[Battaglia et~al., 2018]{battaglia2018relational}
Battaglia, P.~W., Hamrick, J.~B., Bapst, V., Sanchez-Gonzalez, A., Zambaldi,
  V., Malinowski, M., Tacchetti, A., Raposo, D., Santoro, A., Faulkner, R.,
  et~al. (2018).
\newblock Relational inductive biases, deep learning, and graph networks.
\newblock {\em arXiv preprint arXiv:1806.01261}.

\bibitem[Bertasius et~al., 2021]{bertasius2021space}
Bertasius, G., Wang, H., and Torresani, L. (2021).
\newblock Is space-time attention all you need for video understanding?
\newblock {\em arXiv preprint arXiv:2102.05095}.

\bibitem[Biewald, 2020]{wandb}
Biewald, L. (2020).
\newblock Experiment tracking with weights and biases.
\newblock Software available from wandb.com.

\bibitem[Bobick, 1997]{bobick1997movement}
Bobick, A.~F. (1997).
\newblock Movement, activity and action: the role of knowledge in the
  perception of motion.
\newblock {\em Philosophical Transactions of the Royal Society of London.
  Series B: Biological Sciences}, 352(1358):1257--1265.

\bibitem[Carreira and Zisserman, 2017]{carreira2017quo}
Carreira, J. and Zisserman, A. (2017).
\newblock Quo vadis, action recognition? a new model and the kinetics dataset.
\newblock In {\em proceedings of the IEEE Conference on Computer Vision and
  Pattern Recognition}, pages 6299--6308.

\bibitem[Contributors, 2020]{2020mmaction2}
Contributors, M. (2020).
\newblock Openmmlab's next generation video understanding toolbox and
  benchmark.

\bibitem[Crasto et~al., 2019]{crasto2019mars}
Crasto, N., Weinzaepfel, P., Alahari, K., and Schmid, C. (2019).
\newblock Mars: Motion-augmented rgb stream for action recognition.
\newblock In {\em Proceedings of the IEEE/CVF Conference on Computer Vision and
  Pattern Recognition}, pages 7882--7891.

\bibitem[Donahue et~al., 2015]{donahue2015long}
Donahue, J., Anne~Hendricks, L., Guadarrama, S., Rohrbach, M., Venugopalan, S.,
  Saenko, K., and Darrell, T. (2015).
\newblock Long-term recurrent convolutional networks for visual recognition and
  description.
\newblock In {\em Proceedings of the IEEE conference on computer vision and
  pattern recognition}, pages 2625--2634.

\bibitem[Dosovitskiy et~al., 2020]{dosovitskiy2020image}
Dosovitskiy, A., Beyer, L., Kolesnikov, A., Weissenborn, D., Zhai, X.,
  Unterthiner, T., Dehghani, M., Minderer, M., Heigold, G., Gelly, S., et~al.
  (2020).
\newblock An image is worth 16x16 words: Transformers for image recognition at
  scale.
\newblock {\em arXiv preprint arXiv:2010.11929}.

\bibitem[Feichtenhofer et~al., 2019]{feichtenhofer2019slowfast}
Feichtenhofer, C., Fan, H., Malik, J., and He, K. (2019).
\newblock Slowfast networks for video recognition.
\newblock In {\em Proceedings of the IEEE/CVF international conference on
  computer vision}, pages 6202--6211.

\bibitem[Feichtenhofer et~al., 2016]{feichtenhofer2016spatiotemporal}
Feichtenhofer, C., Pinz, A., and Wildes, R.~P. (2016).
\newblock Spatiotemporal residual networks for video action recognition. corr
  abs/1611.02155 (2016).
\newblock {\em arXiv preprint arXiv:1611.02155}.

\bibitem[Feichtenhofer et~al., 2017]{feichtenhofer2017spatiotemporal}
Feichtenhofer, C., Pinz, A., and Wildes, R.~P. (2017).
\newblock Spatiotemporal multiplier networks for video action recognition.
\newblock In {\em Proceedings of the IEEE conference on computer vision and
  pattern recognition}, pages 4768--4777.

\bibitem[Fernando et~al., 2016]{FernandoAl:TPAMI16}
Fernando, B., Gavves, E., Oramas~M., J., Ghodrati, A., and Tuytelaars, T.
  (2016).
\newblock Modeling video evolution for action recognition.
\newblock In {\em TPAMI}.

\bibitem[Ghadiyaram et~al., 2019]{ghadiyaram2019large}
Ghadiyaram, D., Tran, D., and Mahajan, D. (2019).
\newblock Large-scale weakly-supervised pre-training for video action
  recognition.
\newblock In {\em Proceedings of the IEEE/CVF Conference on Computer Vision and
  Pattern Recognition}, pages 12046--12055.

\bibitem[Girdhar and Ramanan, 2020]{girdhar2020acCATERDiagnosticDataset2020}
Girdhar, R. and Ramanan, D. (2020).
\newblock {{CATER}}: {{A}} diagnostic dataset for {{Compositional Actions}} and
  {{TEmporal Reasoning}}.
\newblock {\em arXiv:1910.04744 [cs]}.

\bibitem[Goyal et~al., 2017]{goyal2017accurate}
Goyal, P., Doll{\'a}r, P., Girshick, R., Noordhuis, P., Wesolowski, L., Kyrola,
  A., Tulloch, A., Jia, Y., and He, K. (2017).
\newblock Accurate, large minibatch sgd: Training imagenet in 1 hour.
\newblock {\em arXiv preprint arXiv:1706.02677}.

\bibitem[Hara et~al., 2018]{hara2018can}
Hara, K., Kataoka, H., and Satoh, Y. (2018).
\newblock Can spatiotemporal 3d cnns retrace the history of 2d cnns and
  imagenet?
\newblock In {\em Proceedings of the IEEE conference on Computer Vision and
  Pattern Recognition}, pages 6546--6555.

\bibitem[He et~al., 2019]{he2019stnet}
He, D., Zhou, Z., Gan, C., Li, F., Liu, X., Li, Y., Wang, L., and Wen, S.
  (2019).
\newblock Stnet: Local and global spatial-temporal modeling for action
  recognition.
\newblock In {\em Proceedings of the AAAI Conference on Artificial
  Intelligence}, volume~33, pages 8401--8408.

\bibitem[He et~al., 2015]{he2015delving}
He, K., Zhang, X., Ren, S., and Sun, J. (2015).
\newblock Delving deep into rectifiers: Surpassing human-level performance on
  imagenet classification.
\newblock In {\em Proceedings of the IEEE international conference on computer
  vision}, pages 1026--1034.

\bibitem[He et~al., 2016]{10.1007/978-3-319-49409-8_2}
He, Y., Shirakabe, S., Satoh, Y., and Kataoka, H. (2016).
\newblock Human action recognition without human.
\newblock In Hua, G. and J{\'e}gou, H., editors, {\em Computer Vision -- ECCV
  2016 Workshops}, pages 11--17, Cham. Springer International Publishing.

\bibitem[Hinton et~al., 2012]{hinton2012improving}
Hinton, G.~E., Srivastava, N., Krizhevsky, A., Sutskever, I., and
  Salakhutdinov, R.~R. (2012).
\newblock Improving neural networks by preventing co-adaptation of feature
  detectors.
\newblock {\em arXiv preprint arXiv:1207.0580}.

\bibitem[Holtgraves and Srull, 1990]{holtgraves1990ordered}
Holtgraves, T. and Srull, T.~K. (1990).
\newblock Ordered and unordered retrieval strategies in person memory.
\newblock {\em Journal of Experimental Social Psychology}, 26(1):63--81.

\bibitem[Horn and Schunck, 1981]{horn1981determining}
Horn, B.~K. and Schunck, B.~G. (1981).
\newblock Determining optical flow.
\newblock {\em Artificial intelligence}, 17(1-3):185--203.

\bibitem[Hu et~al., 2018]{hu2018acRelationNetworksObject2018}
Hu, H., Gu, J., Zhang, Z., Dai, J., and Wei, Y. (2018).
\newblock Relation {{Networks}} for {{Object Detection}}.
\newblock {\em arXiv:1711.11575 [cs]}.

\bibitem[Hutchinson and Gadepally,
  2020]{hutchinson2020acVideoActionUnderstanding2020}
Hutchinson, M. and Gadepally, V. (2020).
\newblock Video {{Action Understanding}}: {{A Tutorial}}.
\newblock {\em arXiv:2010.06647 [cs]}.

\bibitem[Jain et~al., 2014]{jain2014university}
Jain, M., van Gemert, J., Snoek, C.~G., et~al. (2014).
\newblock University of amsterdam at thumos challenge 2014.
\newblock {\em ECCV THUMOS Challenge}, 2014.

\bibitem[Ji et~al., 2012]{ji20123d}
Ji, S., Xu, W., Yang, M., and Yu, K. (2012).
\newblock 3d convolutional neural networks for human action recognition.
\newblock {\em IEEE transactions on pattern analysis and machine intelligence},
  35(1):221--231.

\bibitem[Jiang et~al., 2019]{jiang2019stm}
Jiang, B., Wang, M., Gan, W., Wu, W., and Yan, J. (2019).
\newblock Stm: Spatiotemporal and motion encoding for action recognition.
\newblock In {\em Proceedings of the IEEE/CVF International Conference on
  Computer Vision}, pages 2000--2009.

\bibitem[Johnson et~al., 2017]{Johnson_2017_CVPR}
Johnson, J., Hariharan, B., van~der Maaten, L., Fei-Fei, L., Lawrence~Zitnick,
  C., and Girshick, R. (2017).
\newblock Clevr: A diagnostic dataset for compositional language and elementary
  visual reasoning.
\newblock In {\em Proceedings of the IEEE Conference on Computer Vision and
  Pattern Recognition (CVPR)}.

\bibitem[Karpathy et~al., 2014a]{karpathy2014large}
Karpathy, A., Toderici, G., Shetty, S., Leung, T., Sukthankar, R., and Fei-Fei,
  L. (2014a).
\newblock Large-scale video classification with convolutional neural networks.
\newblock In {\em Proceedings of the IEEE conference on Computer Vision and
  Pattern Recognition}, pages 1725--1732.

\bibitem[Karpathy et~al., 2014b]{KarpathyCVPR14}
Karpathy, A., Toderici, G., Shetty, S., Leung, T., Sukthankar, R., and Fei-Fei,
  L. (2014b).
\newblock Large-scale video classification with convolutional neural networks.
\newblock In {\em CVPR}.

\bibitem[Kuehne et~al., 2011]{Kuehne11}
Kuehne, H., Jhuang, H., Garrote, E., Poggio, T., and Serre, T. (2011).
\newblock {HMDB}: a large video database for human motion recognition.
\newblock In {\em Proceedings of the International Conference on Computer
  Vision (ICCV)}.

\bibitem[Lan et~al., 2015]{lan2015beyond}
Lan, Z., Lin, M., Li, X., Hauptmann, A.~G., and Raj, B. (2015).
\newblock Beyond gaussian pyramid: Multi-skip feature stacking for action
  recognition.
\newblock In {\em Proceedings of the IEEE conference on computer vision and
  pattern recognition}, pages 204--212.

\bibitem[Levi and Ullman, 2018]{levi2018efficient}
Levi, H. and Ullman, S. (2018).
\newblock Efficient coarse-to-fine non-local module for the detection of small
  objects.
\newblock {\em arXiv preprint arXiv:1811.12152}.

\bibitem[Luo and Yuille, 2019]{luo2019grouped}
Luo, C. and Yuille, A.~L. (2019).
\newblock Grouped spatial-temporal aggregation for efficient action
  recognition.
\newblock In {\em Proceedings of the IEEE/CVF International Conference on
  Computer Vision}, pages 5512--5521.

\bibitem[Peng et~al., 2014]{peng2014action}
Peng, X., Zou, C., Qiao, Y., and Peng, Q. (2014).
\newblock Action recognition with stacked fisher vectors.
\newblock In {\em European Conference on Computer Vision}, pages 581--595.
  Springer.

\bibitem[Ren et~al., 2015]{ren2015faster}
Ren, S., He, K., Girshick, R., and Sun, J. (2015).
\newblock Faster r-cnn: Towards real-time object detection with region proposal
  networks.
\newblock {\em Advances in neural information processing systems}, 28:91--99.

\bibitem[Rezatofighi et~al., 2017]{rezatofighi2017deepsetnet}
Rezatofighi, S.~H., BG, V.~K., Milan, A., Abbasnejad, E., Dick, A., and Reid,
  I. (2017).
\newblock Deepsetnet: Predicting sets with deep neural networks.
\newblock In {\em 2017 IEEE International Conference on Computer Vision
  (ICCV)}, pages 5257--5266. IEEE.

\bibitem[Santoro et~al., 2017]{santoro2017acSimpleNeuralNetwork2017}
Santoro, A., Raposo, D., Barrett, D. G.~T., Malinowski, M., Pascanu, R.,
  Battaglia, P., and Lillicrap, T. (2017).
\newblock A simple neural network module for relational reasoning.
\newblock {\em arXiv:1706.01427 [cs]}.

\bibitem[Shamsian et~al., 2020]{shamsian2020learning}
Shamsian, A., Kleinfeld, O., Globerson, A., and Chechik, G. (2020).
\newblock Learning object permanence from video.
\newblock In {\em European Conference on Computer Vision}, pages 35--50.
  Springer.

\bibitem[Shanahan et~al., 2020]{shanahan2020acsExplicitlyRelationalNeural2020}
Shanahan, M., Nikiforou, K., Creswell, A., Kaplanis, C., Barrett, D., and
  Garnelo, M. (2020).
\newblock An {{Explicitly Relational Neural Network Architecture}}.
\newblock {\em arXiv:1905.10307 [cs, stat]}.

\bibitem[Shoham, 1987]{shoham1987reasoning}
Shoham, Y. (1987).
\newblock {\em Reasoning about change: time and causation from the standpoint
  of artificial intelligence}.
\newblock PhD thesis, Yale University.

\bibitem[Simonyan and Zisserman, 2014]{simonyan2014two}
Simonyan, K. and Zisserman, A. (2014).
\newblock Two-stream convolutional networks for action recognition in videos.
\newblock {\em arXiv preprint arXiv:1406.2199}.

\bibitem[Soomro et~al., 2012]{soomro2012ucf101}
Soomro, K., Zamir, A.~R., and Shah, M. (2012).
\newblock Ucf101: A dataset of 101 human actions classes from videos in the
  wild.
\newblock {\em arXiv preprint arXiv:1212.0402}.

\bibitem[Stroud et~al., 2020]{stroud2020d3d}
Stroud, J., Ross, D., Sun, C., Deng, J., and Sukthankar, R. (2020).
\newblock D3d: Distilled 3d networks for video action recognition.
\newblock In {\em Proceedings of the IEEE/CVF Winter Conference on Applications
  of Computer Vision}, pages 625--634.

\bibitem[Taylor et~al., 2010]{taylor2010convolutional}
Taylor, G.~W., Fergus, R., LeCun, Y., and Bregler, C. (2010).
\newblock Convolutional learning of spatio-temporal features.
\newblock In {\em European conference on computer vision}, pages 140--153.
  Springer.

\bibitem[Tran et~al., 2015]{tran2015learning}
Tran, D., Bourdev, L., Fergus, R., Torresani, L., and Paluri, M. (2015).
\newblock Learning spatiotemporal features with 3d convolutional networks.
\newblock In {\em Proceedings of the IEEE international conference on computer
  vision}, pages 4489--4497.

\bibitem[Tran et~al., 2017]{tran2017convnet}
Tran, D., Ray, J., Shou, Z., Chang, S.-F., and Paluri, M. (2017).
\newblock Convnet architecture search for spatiotemporal feature learning.
\newblock {\em arXiv preprint arXiv:1708.05038}.

\bibitem[Tran et~al., 2018]{tran2018closer}
Tran, D., Wang, H., Torresani, L., Ray, J., LeCun, Y., and Paluri, M. (2018).
\newblock A closer look at spatiotemporal convolutions for action recognition.
\newblock In {\em Proceedings of the IEEE conference on Computer Vision and
  Pattern Recognition}, pages 6450--6459.

\bibitem[Vaswani et~al., 2017]{vaswani2017attention}
Vaswani, A., Shazeer, N., Parmar, N., Uszkoreit, J., Jones, L., Gomez, A.~N.,
  Kaiser, {\L}., and Polosukhin, I. (2017).
\newblock Attention is all you need.
\newblock In {\em Advances in neural information processing systems}, pages
  5998--6008.

\bibitem[Wang et~al., 2011]{wang:inria-00583818}
Wang, H., Kl{\"a}ser, A., Schmid, C., and Cheng-Lin, L. (2011).
\newblock {Action Recognition by Dense Trajectories}.
\newblock In {\em {CVPR 2011 - IEEE Conference on Computer Vision and Pattern
  Recognition}}, pages 3169--3176, Colorado Springs, United States. {IEEE}.

\bibitem[Wang and Schmid, 2013]{6751553}
Wang, H. and Schmid, C. (2013).
\newblock Action recognition with improved trajectories.
\newblock In {\em 2013 IEEE International Conference on Computer Vision}, pages
  3551--3558.

\bibitem[Wang et~al., 2015]{wang2015towards}
Wang, L., Xiong, Y., Wang, Z., and Qiao, Y. (2015).
\newblock Towards good practices for very deep two-stream convnets.
\newblock {\em arXiv preprint arXiv:1507.02159}.

\bibitem[Wang et~al., 2016]{wang2016temporal}
Wang, L., Xiong, Y., Wang, Z., Qiao, Y., Lin, D., Tang, X., and Van~Gool, L.
  (2016).
\newblock Temporal segment networks: Towards good practices for deep action
  recognition.
\newblock In {\em European conference on computer vision}, pages 20--36.
  Springer.

\bibitem[Wang et~al., 2018]{wang2018non}
Wang, X., Girshick, R., Gupta, A., and He, K. (2018).
\newblock Non-local neural networks.
\newblock In {\em Proceedings of the IEEE conference on computer vision and
  pattern recognition}, pages 7794--7803.

\bibitem[Wu et~al., 2019]{wu2019long}
Wu, C.-Y., Feichtenhofer, C., Fan, H., He, K., Krahenbuhl, P., and Girshick, R.
  (2019).
\newblock Long-term feature banks for detailed video understanding.
\newblock In {\em Proceedings of the IEEE/CVF Conference on Computer Vision and
  Pattern Recognition}, pages 284--293.

\bibitem[Wu et~al., 2020]{wu2020motionnet}
Wu, P., Chen, S., and Metaxas, D.~N. (2020).
\newblock Motionnet: Joint perception and motion prediction for autonomous
  driving based on bird's eye view maps.
\newblock In {\em Proceedings of the IEEE/CVF Conference on Computer Vision and
  Pattern Recognition}, pages 11385--11395.

\bibitem[Yin et~al., 2020]{yin2020disentangled}
Yin, M., Yao, Z., Cao, Y., Li, X., Zhang, Z., Lin, S., and Hu, H. (2020).
\newblock Disentangled non-local neural networks.
\newblock In {\em European Conference on Computer Vision}, pages 191--207.
  Springer.

\bibitem[Yue-Hei~Ng et~al., 2015]{yue2015beyond}
Yue-Hei~Ng, J., Hausknecht, M., Vijayanarasimhan, S., Vinyals, O., Monga, R.,
  and Toderici, G. (2015).
\newblock Beyond short snippets: Deep networks for video classification.
\newblock In {\em Proceedings of the IEEE conference on computer vision and
  pattern recognition}, pages 4694--4702.

\bibitem[Zambaldi et~al., 2018]{zambaldi2018relational}
Zambaldi, V., Raposo, D., Santoro, A., Bapst, V., Li, Y., Babuschkin, I.,
  Tuyls, K., Reichert, D., Lillicrap, T., Lockhart, E., et~al. (2018).
\newblock Relational deep reinforcement learning.
\newblock {\em arXiv preprint arXiv:1806.01830}.

\bibitem[Zhang et~al., 2019]{NEURIPS2019_6e79ed05}
Zhang, Y., Hare, J., and Prugel-Bennett, A. (2019).
\newblock Deep set prediction networks.
\newblock In Wallach, H., Larochelle, H., Beygelzimer, A., d\textquotesingle
  Alch\'{e}-Buc, F., Fox, E., and Garnett, R., editors, {\em Advances in Neural
  Information Processing Systems}, volume~32. Curran Associates, Inc.

\bibitem[Zhou et~al., 2018]{zhou2018temporal}
Zhou, B., Andonian, A., Oliva, A., and Torralba, A. (2018).
\newblock Temporal relational reasoning in videos.
\newblock In {\em Proceedings of the European Conference on Computer Vision
  (ECCV)}, pages 803--818.

\bibitem[Zhu et~al., 2020]{zhu2020acComprehensiveStudyDeep2020}
Zhu, Y., Li, X., Liu, C., Zolfaghari, M., Xiong, Y., Wu, C., Zhang, Z., Tighe,
  J., Manmatha, R., and Li, M. (2020).
\newblock A {{Comprehensive Study}} of {{Deep Video Action Recognition}}.
\newblock {\em arXiv:2012.06567 [cs]}.

\end{thebibliography}
